\documentclass[11pt]{article}

\usepackage{acl}
\usepackage{times}
\usepackage{latexsym}

\usepackage[T1]{fontenc}

\usepackage[utf8]{inputenc}

\usepackage{microtype}

\usepackage{inconsolata}

\usepackage{graphicx}

\usepackage{amsmath}
\usepackage{amssymb}
\usepackage{booktabs}
\usepackage{multirow}

\title{Listwise Direct Preference Optimization with Multi-Dimensional Preference Mixing}

\author{
Yuhui Sun \\
University of Alberta 
\And
Xiyao Wang \\
University of Toronto 
\And
Zixi Li \\
Zhejiang University 
\AND
YiTian Ding \\
School of Computer Science \\
McGill University 
\And
Tianyang Ling \\
University of Alberta 
\And
Jialuo Chen \\
University of Toronto 
\AND
Tianyi Yu \\
Alibaba Group (Ant Group) 
\And
Zhenlong Yuan \\
Chinese Academy of Sciences 
\And
Jinman Zhao \\
University of Toronto \\
\texttt{jzhao@cs.toronto.edu}
}

\begin{document}

\maketitle

\begin{abstract}
Recent alignment methods based on Direct Preference Optimization (DPO) reformulate preference learning as supervised optimization over pairwise comparisons, offering improved efficiency and stability over reinforcement learning from human feedback (RLHF). However, existing DPO-style methods implicitly assume a single fixed preference objective, which limits their ability to model the structured and sometimes conflicting nature of real-world human judgments that span multiple preference dimensions. In this work, we propose Listwise Direct Preference Optimization ($\lambda$-DPO), a unified framework that simultaneously improves supervision granularity and preference flexibility. Instead of collapsing multi-dimensional preference signals into a single ranking, $\lambda$-DPO constructs a mixture of listwise preference distributions weighted by a preference vector $\lambda$ on the probability simplex, enabling a single model to internalize a continuous spectrum of preference trade-offs. To further improve robustness, we introduce a performance-driven stochastic $\lambda$ scheduler that adaptively samples preference weights based on empirical downstream performance, explicitly mitigating the risks of misspecification inherent to static weighting schemes. We evaluate our method across multiple model families and scales on six widely used benchmarks. Experimental results show the consistent improvement against baselines.
\end{abstract}

\section{Introduction}
Large language models (LLMs) have demonstrated strong generalization capabilities such as question instruction following~\cite{brown2020language,touvron2023llamaopenefficientfoundation,openai2024gpt4technicalreport}. Despite these advances, their behavior in real-world applications frequently remains misaligned with human expectations, manifesting in issues such as factual inaccuracies, unsafe responses, and failures to appropriately interpret user intent~\cite{bai2022training,ouyang2022training,lin-etal-2022-truthfulqa}. Unlike early formulations that treated alignment as the optimization of a single performance metric, practical alignment objectives are inherently multi-dimensional, requiring models to simultaneously balance factors such as helpfulness, factuality, safety, fluency, and task-specific constraints often with intrinsic tensions and trade-offs among them~\cite{christiano2017deep,wu2021recursively,korbak2023pretraining}. Such human preferences are difficult to compress into a single scalar reward or fixed objective function; instead, they are more naturally expressed through relative comparisons and rankings over multiple candidate responses~\cite{bradley1952rank,burges2005learning,cui2023ultrafeedback}. Consequently, the central challenge of alignment lies not in learning a single static “optimal” answer, but in modeling and generalizing the structured manifestation of multi-dimensional human preferences over sets of candidate outputs.

Multiple methods have gradually emerged as a dominant paradigm for aligning LLMs with human expectations. Among them, reinforcement learning from human feedback (RLHF) optimizes generation policies using preference signals, typically via policy gradient methods such as Proximal Policy Optimization (PPO)~\cite{schulman2017proximalpolicyoptimizationalgorithms} has demonstrated substantial improvements~\cite{ouyang2022training}. However, RLHF often suffers from high engineering and computational overhead, as well as instability arising from reward modeling errors and reinforcement learning dynamics. Direct Preference Optimization (DPO)~\cite{rafailov2023direct} reformulates alignment as a supervised learning problem, directly optimizing the policy using pairwise preference comparisons. Building upon the DPO framework, a growing body of work has proposed various extensions and variants such as CPO~\cite{xu2024contrastive}, SimPO~\cite{meng2024simpo}, RRHF~\cite{yuan2023rrhf}, ORPO~\cite{hong-etal-2024-orpo} and IPO~\cite{azar2024general}. These work modify the optimization objective, regularization strategy, or preference formulation to improve robustness and efficiency under different training regimes.

Existing preference-alignment methods typically rely on an implicit modeling assumption that human preferences induce a single, fixed ranking over candidate responses. Recently, several listwise extensions of DPO such as LiPO~\cite{liu-etal-2025-lipo}, Soft Preference Optimization~\cite{sharifnassab2024softpreferenceoptimizationaligning} have been proposed to directly optimize ranking distributions over sets of candidate responses, primarily aiming to improve supervision density and training stability. However, these methods still fundamentally assume that, for a given prompt, candidate responses correspond to a semantically fixed ranking target. In practice, this assumption is typically implemented by collapsing multi-dimensional preference signals into a single ranking objective. Nevertheless, real-world human preference data do not always support this formulation. For example, in multi-dimensional preference datasets such as UltraFeedback, candidate responses to the same prompt are evaluated independently along multiple criteria which often yield conflicting judgments across dimensions~\cite{cui2023ultrafeedback}. 

Based on the above observations, we propose multi-preference weighted listwise DPO \textbf{(\(\lambda\)-DPO)}, a unified framework that simultaneously improves supervision granularity and preference flexibility. Specifically, instead of simplifying preference signals into binary pairwise comparisons, our method leverages the multi-candidate ranking information available for each prompt and directly learns to match a \emph{listwise preference distribution} with the model-induced distribution. This formulation provides richer and lower-variance training signals while capturing global ranking relationships over candidate sets. To account for the inherently multi-dimensional nature of real-world alignment objectives, we further introduce a preference weight vector \(\lambda\) defined on the probability simplex, which combines listwise preference distributions corresponding to different preference dimensions via a weighted mixture, yielding a controllable target distribution \(p_{\lambda}\). This design allows a single model to internalize a continuous spectrum of preference trade-offs during training and enables inference-time controllable alignment through adjusting \(\lambda\) without retraining. In addition, we design a \emph{learned \(\lambda\) scheduling strategy} that dynamically adapts the sampling distribution of \(\lambda\) based on empirical performance, improving training robustness and generalization while remaining compatible with resource-constrained settings. In summary, this paper makes the following contributions:
\begin{itemize}
    \item We identify a fundamental limitation of existing DPO and its extensions, which rely on binary supervision and a single fixed alignment objective, making them insufficient to capture the structure of multi-dimensional human preferences. To address this, we propose a multi-preference weighted listwise DPO framework.
    \item We empirically validate the effectiveness of \(\lambda\)-DPO across multiple model scales and evaluation benchmarks, demonstrating consistent improvements over existing DPO settings.
    \item Through ablation and analysis experiments, we systematically study the effects of different preference weight configurations and the learned \(\lambda\) scheduling strategy, shedding light on how preference weighting influences training stability and alignment performance.
\end{itemize}

\begin{figure*}[h]
    \centering
    \includegraphics[width=\linewidth]{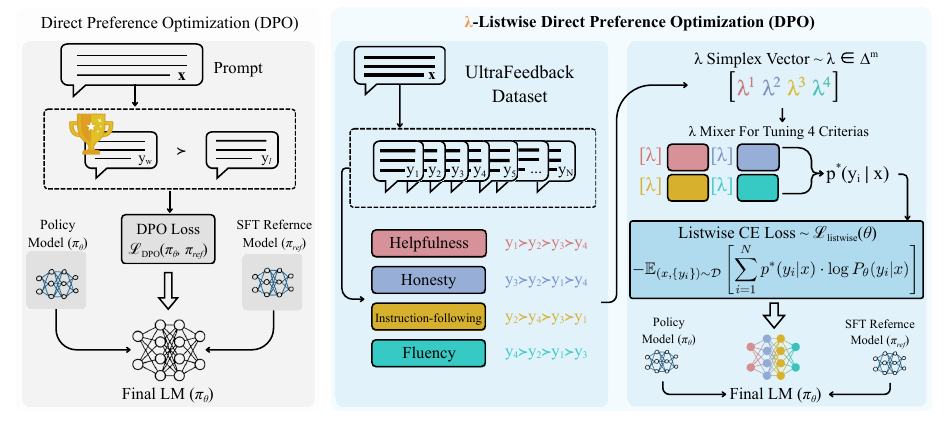}
    \caption{Overall framework.}
    \label{framework}
\end{figure*}

\section{Related Work}
\paragraph{Alignment and RLHF.} RLHF~\cite{ouyang2022training} is a foundational paradigm for aligning LLMs. Its core approach learns a reward model from human preference comparisons and optimizes the generation policy using policy gradient methods such as REINFORCE~\cite{sutton1999policy}, PPO~\cite{schulman2017proximalpolicyoptimizationalgorithms} or REINFORCE++~\cite{hu2025reinforcestabilizingcriticfreepolicy}. While effective, RLHF typically incurs high computational cost and suffers from training instability. To address these limitations, recent work explores alternatives that bypass explicit value or reward modeling, such as GRPO~\cite{shao2024deepseekmathpushinglimitsmathematical}, or reformulate alignment as a supervised learning problem, such as DPO~\cite{rafailov2023direct}.

\paragraph{DPO Variation.} DPO has inspired a growing family of alignment methods that refine its objective or optimization strategy. Representative variants include SimPO~\cite{meng2024simpo} simplifies the DPO objective by removing explicit reference policies; ORPO~\cite{hong-etal-2024-orpo} unifies preference optimization with likelihood training; CTO~\cite{xu2024contrastive} and IPO~\cite{azar2024general} reinterpret preference learning from contrastive or implicit policy optimization perspectives; StepDPO~\cite{lai2024stepdpostepwisepreferenceoptimization,xu-etal-2025-full} extends DPO to step-level supervision for multi-step reasoning; and RDPO~\cite{park-etal-2024-disentangling} explicitly disentangles response length from quality to mitigate length bias in preference optimization. More recently, online DPO methods study in interactive or streaming settings, highlighting the crucial role of sampling strategies in stabilizing and improving online alignment performance~\cite{shi2025the,tu2025enhancing,pang2024iterative}. Despite differences in formulation and optimization details, these methods largely share a common reliance on pairwise supervision and assume a single implicit alignment objective. Recent surveys~\cite{xiao2025comprehensivesurveydirectpreference} provide systematic overviews of DPO-style methods, highlighting both their efficiency advantages and shared modeling assumptions.

\paragraph{Listwise Preference Optimization.} Recent work has explored preference optimization from pairwise comparisons to \emph{listwise} feedback, aiming to more fully exploit the relative information among multiple candidate responses for a given prompt. Representative approaches include LiPO~\cite{liu-etal-2025-lipo} and Soft Preference Optimization~\cite{sharifnassab2024softpreferenceoptimizationaligning} formulate alignment as directly matching model distributions to target ranking distributions over response lists. Ordinal Preference Optimization~\cite{zhao-etal-2025-permutative} optimizes differentiable ranking metrics, and listwise extensions of DPO that leverage soft rankings or best-of-$K$ feedback. Beyond language models, listwise preference optimization has also been applied in other domains, such as vision–language models~\cite{pesaran-zadeh-etal-2025-lpoi} and diffusion-based generative models~\cite{bai2025betteroptimizationlistwisepreference}.

\section{Preliminaries}
We briefly review preference modeling and DPO. These form the foundation of our approach. 
\paragraph{Preference Modeling.}
Let $x$ denote a prompt, and $y_w \succ y_l$ a human preference indicating that
completion $y_w$ is preferred over $y_l$.
We assume such comparisons are drawn from an underlying preference distribution
$p^*(y_w \succ y_l \mid x)$.
A common instantiation is the Bradley--Terry model~\cite{bradley1952rank} that parameterizes preferences via a latent scoring function $r(x,y)$:
\begin{align}
   & p^*(y_w \succ y_l \mid x)\nonumber\\ =& \frac{\exp(r(x, y_w))}{\exp(r(x, y_w)) + \exp(r(x, y_l))} \nonumber  \\
    =& \sigma\!\left(r(x,y_w) - r(x,y_l)\right) 
\end{align}

\paragraph{KL-Regularized RL Objective.} 
Once a reward model is trained, the standard RLHF approach fine-tunes the policy \( \pi_\theta \) to maximize expected rewards while constraining deviation from the reference policy \( \pi_{\text{ref}} \) via KL regularization:
\begin{align}
\label{eq:kl-rl}
\max_{\pi_\theta} \mathbb{E}_{x \sim \mathcal{D}, y \sim \pi_\theta(y|x)}  & \big[  r(x, y) \\
& - \beta \mathrm{KL}\left( \pi_\theta(y | x) \| \pi_{\text{ref}}(y | x) \right) \big] \nonumber
\end{align}
where \( \beta > 0 \) controls the strength of the KL constraint. This objective encourages the policy to produce high-reward outputs while staying close to \( \pi_{\text{ref}} \). However, in practice, it requires sampling, reward evaluation, and reinforcement learning algorithms such as PPO~\cite{schulman2017proximalpolicyoptimizationalgorithms} to optimize.

\paragraph{Reparameterization and DPO.} 
DPO sidesteps explicit reward modeling and reinforcement learning by recognizing that the optimal policy \( \pi_r \) under Equation~\ref{eq:kl-rl} can be written as:
\begin{equation}
\pi_r(y|x) \propto \pi_{\text{ref}}(y|x) \cdot \exp\left( \frac{1}{\beta} r(x, y) \right)
\label{eq:pi-r}
\end{equation}
Using Equation~\ref{eq:pi-r}, we reparameterize the reward difference into the log-ratio of policies:
\begin{align}
& \beta(r(x, y_w) - r(x, y_l)) \nonumber\\
&= \log \frac{\pi_r(y_w|x)}{\pi_{\text{ref}}(y_w|x)} - \log \frac{\pi_r(y_l|x)}{\pi_{\text{ref}}(y_l|x)}
\end{align}
Replacing \( \pi_r \) with \( \pi_\theta \) yields the DPO loss:
{

\begin{align}
\label{eq:dpo-loss}
- \mathbb{E}_{(x, y_w, y_l) \sim \mathcal{D}} \bigg[ 
\log \sigma
& \bigg( 
\beta \log \frac{\pi_\theta(y_w|x)}{\pi_{\text{ref}}(y_w|x)} \\
& - \beta \log \frac{\pi_\theta(y_l|x)}{\pi_{\text{ref}}(y_l|x)} 
\bigg) 
\bigg] \nonumber
\end{align}
}

This loss directly optimizes the policy parameters \( \theta \) using preference comparisons, without sampling or reinforcement learning. It forms the foundation upon which we build our multi-preference and listwise extensions in later sections.

\section{Multi-Preference Lambda-weighted Listwise DPO}
Motivated by the limitations of pairwise DPO under a single fixed preference distribution, we propose two orthogonal enhancements: listwise supervision and simplex-weighted optimization. First, instead of relying on binary comparisons, we adopt listwise learning to utilize full human rankings over multiple outputs. This provides stronger supervision signals and captures richer preference structure. Second, we generalize DPO to support multi-objective alignment by introducing a simplex-weighted aggregation over multiple reward components. The resulting $\lambda$-DPO formulation enables flexible interpolation across alignment dimensions such as helpfulness, harmlessness, and fluency. Together, these extensions allow controllable and scalable alignment with more nuanced human feedback. The overview pipeline is illustrated in Figure~\ref{framework}.

\subsection{Listwise Preference Modeling}
Let \( \{y_1, y_2, \dots, y_N\} \) be a set of candidate completions for a prompt \( x \), and let \( p^*(y_i | x) \) be a listwise human preference distribution over these candidates. Analogous to the pairwise DPO reparameterization, we define the listwise policy distribution under \( \pi_\theta \) as:
\begin{equation}
P_\theta(y_i | x) = \frac{\left( \frac{\pi_\theta(y_i|x)}{\pi_{\text{ref}}(y_i|x)} \right)^\beta}{\sum_{j=1}^{N} \left( \frac{\pi_\theta(y_j|x)}{\pi_{\text{ref}}(y_j|x)} \right)^\beta}
\end{equation}
which defines a softmax over reweighted log-ratios, consistent with KL-derived optimal policies.

We then define the listwise DPO objective as a cross-entropy loss between human preferences \( p^*(y_i|x) \) and model predictions:
\begin{align}
& \mathcal{L}_{\text{listwise}}(\theta) =  \\
& -\mathbb{E}_{(x, \{y_i\}) \sim \mathcal{D}}
\left[ \sum_{i=1}^{N} p^*(y_i | x) \cdot \log P_\theta(y_i | x) \right] \nonumber
\end{align}

\subsection{Lambda-weighted Preference Composition}
To enable multi-objective control, we assume access to \( m \) distinct preference dimensions, each defined by a distribution \( p^{*(k)}(y_i|x) \) for objective \( k \in \{1, \dots, m\} \). For each prompt, we sample or specify a weight vector \( \lambda \in \Delta^m \) (the probability simplex), and define the aggregated preference distribution:
\begin{equation}
p^\lambda(y_i | x) = \sum_{k=1}^m \lambda_k \cdot p^{*(k)}(y_i | x)
\end{equation}
This allows us to interpolate among various alignment goals without retraining.

The full Lambda-weighted Listwise DPO objective becomes:
\begin{align}
& \mathcal{L}_{\lambda\text{-DPO}}(\theta) = \\
& -\mathbb{E}_{(x, \{y_i\}), \lambda} \left[ \sum_{i=1}^{N} p^\lambda(y_i | x) \cdot \log P_\theta(y_i | x) \right] \nonumber
\end{align}
This expectation can be approximated by sampling \( \lambda \sim \text{Unif}(\Delta^m) \) or conditioning on user-specified preferences.

\subsection{What does $\lambda$-DPO update do?}
To understand the optimization dynamics, consider the gradient of the loss:
{
\begin{align}
\nabla_\theta \mathcal{L}_{\lambda\text{-DPO}} = - \mathbb{E}_{x, \lambda} \bigg[ & \sum_{i} \left( p^\lambda(y_i | x) - P_\theta(y_i | x) \right) \nonumber \\
&\ \  \cdot \nabla_\theta \log \pi_\theta(y_i | x) \bigg]
\end{align}
}
This gradient updates the model to increase likelihood of responses aligned with weighted human preferences, while decreasing likelihood of responses overrepresented in the reference policy. Importantly, when \( \lambda \) is fixed (e.g., \( \lambda = (1, 0, 0) \)), the objective reduces to a single-objective listwise DPO loss. Due to space limitations, full derivations of the $\lambda$-DPO objective and its gradient are deferred to Appendix~\ref{sec:derivation}.

\subsection{The Pipeline}
The training process proceeds as follows:
\begin{enumerate}
  \item For each prompt \( x \), obtain \( N \) candidate completions \( \{y_1, \dots, y_N\} \sim \pi_{\text{ref}}(\cdot|x) \).
  \item For each alignment dimension \( k \), elicit a human listwise preference distribution \( p^{*(k)}(y_i | x) \).
  \item Sample or define \( \lambda \in \Delta^m \), and compute the mixture preference \( p^\lambda \).
  \item Optimize \( \pi_\theta \) using the listwise cross-entropy loss between \( p^\lambda \) and \( P_\theta \).
\end{enumerate}
In practice, we can reuse datasets where multiple preference criteria are annotated. During inference, changing \( \lambda \) enables personalized alignment without modifying model parameters.
\begin{table*}[t]
\centering
\resizebox{\textwidth}{!}{
\begin{tabular}{l|l|cccccc|c}
\toprule
\textbf{Model} & \textbf{Method} & \textbf{MMLU} & \textbf{ARC} & \textbf{HellaSwag} & \textbf{TruthfulQA} & \textbf{WSC273} & \textbf{MathQA} & \textbf{Average} \\
\midrule

\multirow{3}{*}{Llama 3.2-1B Instruct}       
& SIMPO         & 47.52 & 39.75 & 46.06 & 26.07 & 71.79 & 30.72 & 43.65 \\
& DPO           & 47.52 & 40.37 & 45.87 & 26.07 & 71.43 & 30.89 & 43.69 \\
& $\lambda$-DPO & \textbf{48.71} & \textbf{45.61} & \textbf{46.15} & \textbf{29.01} & \textbf{75.09} & \textbf{32.83} & \textbf{46.23} \\
\midrule

\multirow{3}{*}{Qwen 1.8B Chat}           
& SIMPO         & 44.29 & 32.27 & 43.11 & 15.91 & 60.81 & 24.56 & 36.83 \\
& DPO           & 44.30 & 32.72 & 43.17 & 15.91 & 61.54 & 23.38 & 36.84 \\
& $\lambda$-DPO & \textbf{45.01} & \textbf{38.51} & \textbf{43.82} & \textbf{18.12} & \textbf{64.84} & \textbf{29.41} & \textbf{39.95} \\
\midrule

\multirow{3}{*}{Llama 3-8B Instruct}           
& SIMPO         & \textbf{63.21} & 59.97 & \textbf{77.45} & 45.17 & 77.66 & 31.93 & 59.23 \\
& DPO           & 61.49 & 63.16 & 76.33 & 45.53 & 72.53 & 31.59 & 58.44 \\
& $\lambda$-DPO & 61.03 & \textbf{73.25} & 76.44 & \textbf{46.63} & \textbf{85.35} & \textbf{36.87} & \textbf{63.26} \\
\midrule

\multirow{3}{*}{Qwen2 7B Instruct}           
& SIMPO         & \textbf{68.58} & 68.65 & 62.63 & 45.17 & 76.92 & 39.80 & 60.29 \\
& DPO           & 68.11 & \textbf{69.50} & 62.50 & 45.41 & 75.09 & 42.53 & 60.52 \\
& $\lambda$-DPO & 68.17 & 69.22 & \textbf{63.60} & \textbf{46.27} & \textbf{78.75} & \textbf{42.91} & \textbf{61.49} \\
\bottomrule
\end{tabular}}

\caption{
Accuracy (\%) comparison across six widely used benchmarks for different preference
optimization methods and model backbones.
Each block corresponds to a specific model, and rows compare SIMPO, DPO, and
$\lambda$-DPO.
Columns report performance on MMLU, ARC, HellaSwag, TruthfulQA, WSC273, and MathQA,
along with the averaged score across all benchmarks.
Bold values indicate the best performance within each model group.
}
\label{tab:method-comparison-full}
\end{table*}
\section{Theoretical Analysis}
While DPO offers a principled and scalable approach to aligning language models with human feedback, its reliance on pairwise supervision imposes intrinsic limitations in expressive power, particularly in multi-objective alignment settings. In this section, we provide a theoretical analysis of this limitation and motivate our proposed extensions for better capturing and aggregating multi-dimensional human preferences.

\subsection{Pairwise Limitation}
Given a prompt \(x\) and a sampled pair of completions \(y_w\) (preferred) and
\(y_l\) (dispreferred), Direct Preference Optimization (DPO) relies on binary
preference supervision.
From a listwise perspective, such supervision can be interpreted as inducing an
implicit target distribution over the sampled pair:
\begin{equation}
p^*(y_i \mid x) =
\begin{cases}
1 & \text{if } y_i = y_w, \\
0 & \text{if } y_i = y_l.
\end{cases}
\label{eq:pairwise-target}
\end{equation}
This formulation is simple and computationally efficient, allowing policy
optimization directly from binary human feedback.

However, treating preferences as hard assignments discards finer-grained signals,
such as partial preference strength or equivalence between responses.
Moreover, because gradient updates depend on a single sampled pair, the resulting
optimization can exhibit high variance across iterations.
Since the objective only constrains the relative preference between \(y_w\) and
\(y_l\), other plausible or high-quality completions for the same prompt are
ignored.
Consequently, the expressive capacity of pairwise DPO is limited when multiple
responses are comparably suitable, a limitation that becomes particularly
pronounced in multi-objective alignment settings.

\subsection{Pairwise Supervision as a Degenerate Listwise Objective}

Equation~\eqref{eq:pairwise-target} can be viewed as a degenerate form of listwise
preference supervision, where the target distribution assigns all probability
mass to a single completion within a sampled pair.
More generally, let $\mathcal{Y}(x)=\{y_1,\dots,y_K\}$ denote a set of candidate
completions for prompt $x$.
A listwise preference objective aims to model a target distribution
$p^*(\cdot \mid x)$ over $\mathcal{Y}(x)$, satisfying
\begin{equation}
\sum_{i=1}^K p^*(y_i \mid x) = 1, \qquad p^*(y_i \mid x) \ge 0.
\label{eq:listwise-simplex}
\end{equation}

Under pairwise DPO, the effective target distribution collapses to a distribution
supported on only two sampled completions, with all mass assigned to the preferred
one.
This corresponds to an extreme point of the probability simplex in
Eq.~\eqref{eq:listwise-simplex}, with zero entropy.
As a result, the optimization objective is insensitive to the relative quality of
other candidates in $\mathcal{Y}(x)$, even when multiple responses are
comparably suitable under certain preference trade-offs.

\subsection{Incompatibility with Multi-Objective Preferences}

Consider a setting with multiple preference dimensions, such as helpfulness,
factuality, and safety.
Let $r^{(k)}(x,y)$ denote a latent scoring function associated with the $k$-th
preference criterion.
Each criterion induces its own ordering over candidate responses in
$\mathcal{Y}(x)$.
In general, there does not exist a single scalar function $r(x,y)$ such that
\begin{align}
& r(x,y_i) > r(x,y_j) \\
& \quad \Longleftrightarrow \quad 
r^{(k)}(x,y_i) > r^{(k)}(x,y_j) \nonumber 
\;\; \forall k.
\label{eq:no-single-ordering}
\end{align}

As a consequence, different preference dimensions may induce conflicting
pairwise comparisons over the same response set.
Reducing such multi-dimensional preference signals to a single binary comparison
necessarily discards information about preference trade-offs.
From this perspective, the limitation of pairwise DPO is structural: binary
comparisons cannot represent preference distributions that depend on multiple,
potentially competing objectives.

\begin{table*}[t]
\centering
\resizebox{\textwidth}{!}{
\begin{tabular}{l|ccccccc}
\toprule
\textbf{$\lambda$ Setting} & \textbf{MMLU} & \textbf{ARC} & \textbf{HellaSwag} & \textbf{TruthfulQA} & \textbf{WSC273} & \textbf{MathQA} & \textbf{Average} \\
\midrule

$[1, 0, 0, 0]$ (Helpfulness) 
& 46.36 & 48.91 & 46.22 & 28.03 & 70.70 & 33.53 & 45.63 \\

$[0, 1, 0, 0]$ (Honesty) 
& 46.48 & 49.04 & \textbf{46.25} & 27.91 & 70.33 & 33.63 & 45.61 \\

$[0, 0, 1, 0]$ (Instruction-Following) 
& 46.39 & \textbf{49.29} & 46.24 & 28.03 & 71.06 & \textbf{33.67} & 45.78 \\

$[0, 0, 0, 1]$ (Fluency) 
& 46.26 & 48.84 & 46.22 & 28.15 & 70.33 & 33.37 & 45.53 \\

$[0.25, 0.25, 0.25, 0.25]$ (Uniform) 
& \textbf{48.71} & 45.61 & 46.15 & \textbf{29.01} & \textbf{75.09} & 32.83 & \textbf{46.23} \\


\bottomrule
\end{tabular}}

\caption{
Ablation study of different $\lambda$-weight configurations in $\lambda$-DPO,
conducted on the \textit{LLaMA 3.2-1B Instruct}.
}
\label{tab:lambda-ablation}
\end{table*}
\section{Results}
\subsection{Experimental Setup}
\paragraph{Model Selection.} We evaluate our approach on representative open-weight LLMs from both the Llama3~\cite{touvron2023llamaopenefficientfoundation} and Qwen~\cite{qwen2025qwen25technicalreport} families. Specifically, we consider \textit{Llama 3.2-1B Instruct}, \textit{Llama 3-8B Instruct}, \textit{Qwen 1.8B Chat} and \textit{Qwen2 7B Instruct}.
\paragraph{Benchmark Selection.} To evaluate the effectiveness of our approach across diverse reasoning and alignment-sensitive tasks, we conduct experiments on a collection of widely used
benchmarks spanning knowledge-intensive reasoning, commonsense inference, factual
accuracy, and mathematical problem solving.
Specifically, we include MMLU~\cite{hendrycks2021measuring}, ARC~\cite{clark2018think},
HellaSwag~\cite{zellers-etal-2019-hellaswag}, TruthfulQA~\cite{lin-etal-2022-truthfulqa},
WSC273~\cite{levesque2012winograd}, and MathQA~\cite{amini-etal-2019-mathqa}. Detailed benchmarks statistics reported in
Appendix~\ref{sec: bench}.
\paragraph{Baseline Selection.} We compare our method against representative preference-based alignment baselines that are closely related to our formulation. In particular, we include DPO~\cite{rafailov2023direct} and SimPO~\cite{meng2024simpo}.
\paragraph{Training Detail.} We set the preference scaling parameter $\beta$ to 0.1, following common practice in prior work. Models are trained for three epochs with a learning rate of $5\times10^{-6}$, using a cosine learning rate scheduler with a warmup ratio of 0.1. We use an batch size of 32, and the maximum sequence length is set to 2048 tokens. Our experiments are conducted on 4xH200 GPUs.
\paragraph{Dataset.} We conduct fine-tuning using the Ultrafeedback dataset~\cite{cui2023ultrafeedback},
a large-scale, instruction-following corpus designed for alignment and preference
optimization. Ultrafeedback contains diverse prompts and multiple candidate responses annotated with
multi-preference signals.
\subsection{Main Results}
We report the main experimental results in Table~\ref{tab:method-comparison-full} across six widely used benchmarks. All results are reported in terms of accuracy (\%), with higher values indicating
better performance. Overall, $\lambda$-DPO consistently achieves the strongest average performance
across all evaluated models, outperforming both SIMPO and DPO in every setting.
The gains are particularly pronounced on benchmarks that are sensitive to
alignment-related properties, such as TruthfulQA and WSC273, where $\lambda$-DPO
yields clear and consistent improvements.
This suggests that moving beyond pairwise supervision and explicitly aggregating
preference signals leads to more effective alignment with nuanced human judgments.

We further observe that $\lambda$-DPO delivers robust improvements across model
scales. For smaller models (e.g., \textit{Llama 3.2-1B Instruct} and \textit{Qwen 1.8B Chat}), $\lambda$-DPO provides
substantial relative gains, indicating improved data efficiency under limited
capacity. For larger models (e.g., \textit{Llama 3-8B Instruct} and \textit{Qwen2-7B Instruct}), $\lambda$-DPO continues to
outperform pairwise baselines, demonstrating that the proposed approach scales
effectively with model size. Taken together, these results validate the effectiveness and generality of
$\lambda$-DPO across diverse benchmarks and model families.

A closer inspection of Table~\ref{tab:method-comparison-full} shows that the
performance gains of $\lambda$-DPO are most pronounced on benchmarks that are
sensitive to alignment-related properties, particularly \textit{TruthfulQA} and
\textit{WSC273}. On \textit{WSC273}, $\lambda$-DPO outperforms DPO by
+3.66, +3.30, +12.82, and +3.66 points for \textit{Llama 3.2-1B},
\textit{Qwen 1.8B}, \textit{Llama 3-8B}, and \textit{Qwen2-7B}, respectively,
and also surpasses SIMPO by +3.30, +4.03, +7.69, and +1.83 points on the same
models. Similarly, on \textit{TruthfulQA}, $\lambda$-DPO improves over DPO by
+2.94 (\textit{Llama 3.2-1B}), +2.21 (\textit{Qwen 1.8B}), +1.10
(\textit{Llama 3-8B}), and +0.86 (\textit{Qwen2-7B}), while exceeding SIMPO by
+2.94, +2.21, +1.46, and +1.10 points, indicating more truthful and
human-consistent responses across model scales. The especially large gain on
\textit{WSC273} for \textit{Llama 3-8B} highlights that aggregating preference
signals beyond pairwise supervision can substantially improve alignment-critical
commonsense judgment. In contrast, performance differences on other benchmarks,
including \textit{ARC}, \textit{MathQA}, and the broader commonsense task
\textit{HellaSwag}, are generally more moderate, suggesting that the primary
advantage of $\lambda$-DPO lies in better capturing nuanced human judgments
without sacrificing generalization.

\begin{table*}[t]
\centering
\resizebox{\textwidth}{!}{
\begin{tabular}{l|ccccccc}
\toprule
\textbf{$\lambda$ Setting} & \textbf{MMLU} & \textbf{ARC} & \textbf{HellaSwag} & \textbf{TruthfulQA} & \textbf{WSC273} & \textbf{MathQA} & \textbf{Average} \\
\midrule

Uniform
& 48.71 & 45.61 & 46.15 & 29.01 & 75.09 & 32.83 & 46.23 \\

Learned Scheduler
& 48.35 & 48.76 & 46.20 & 29.03 & 75.06 & 33.53 & \textbf{46.82} \\

\bottomrule
\end{tabular}}

\caption{
Uniform vs. Learned $\lambda$,
conducted on the \textit{LLaMA 3.2-1B Instruct}.
}
\label{tab:lambda-ablation2}
\end{table*}

\subsection{Additional Analysis}

\paragraph{Uniform vs. Unimodal Preference Weights.}
We investigate the effect of different $\lambda$ configurations by comparing a uniform mixture of preference dimensions with unimodal (one-hot) alternatives, and the model is trained to optimize a single objective at a time. Table~\ref{tab:lambda-ablation} reports results on \textit{LLaMA 3.2-1B Instruct}. Across unimodal settings, performance is relatively similar, and no single preference dimension consistently dominates across all benchmarks.
This suggests that the gains of $\lambda$-DPO are not driven by optimizing any
individual objective in isolation. In contrast, the uniform mixture $\lambda=(0.25,0.25,0.25,0.25)$ achieves the best overall average performance, yielding improvements on alignment-sensitive tasks such as TruthfulQA and WSC273 while maintaining competitive accuracy on other
benchmarks. Although certain unimodal configurations achieve slightly higher scores on individual tasks (e.g., instruction-following on ARC), they fail to generalize consistently across the benchmark suite. These results support our central hypothesis that explicitly aggregating multi-dimensional preference signals leads to more robust and balanced alignment than optimizing a single preference dimension alone.

\paragraph{Uniform vs. Learned $\lambda$.}
Static uniform $\lambda$ configurations face two key limitations. First, different benchmarks exhibit different sensitivities to preference dimensions, whereas a uniform $\lambda$ implicitly assumes that all objectives are equally important. This mismatch prevents the model from adapting to benchmark-specific preference structure and can lead to suboptimal performance. Second, although we fit a polynomial performance model $f(\lambda)$ to approximate downstream accuracy over the simplex, this approximation is inherently imperfect: $\lambda$ values predicted as weak by $f$ may still yield strong true performance. Thus, selecting only the maximizer of $f$ risks collapsing to a poorly explored region of the preference space.

Instead of committing to a single $\lambda$, we construct a stochastic scheduler that samples preference weights in a performance-aware manner. Specifically, we define an idealized continuous sampling distribution over the simplex
\[
p(\lambda)\;\propto\;\exp(\tau f(\lambda)), \qquad \lambda\in\Delta_{4},
\]
where $\tau>0$ controls the exploration--exploitation balance. Since this density cannot be computed in closed form, we discretize it by sampling candidate $\{\lambda^{(j)}\}_{j=1}^{k}$ from a Dirichlet prior and applying a softmax:
\[
p(\lambda^{(j)})=
\frac{\exp(\tau f(\lambda^{(j)}))}
{\sum_{l}\exp(\tau f(\lambda^{(l)})) }.
\]
This produces a learned $\lambda$ scheduler that prioritizes promising trade-offs while still exploring alternative preference allocations, explicitly guarding against misspecification of $f$.

The result is presented in Table~\ref{tab:lambda-ablation2}, with the full formulation and implementation details of the learned
$\lambda$ scheduler provided in Appendix~\ref{sec:scheduler}. Empirically, this design yields clear gains. Compared to the uniform mixture $\lambda=(0.25,0.25,0.25,0.25)$, which achieves an average score of $46.23$, the learned scheduler improves the overall average to $46.82$. It further delivers meaningful benchmark-specific improvements, such as ARC ($45.61 \rightarrow 48.76$) and MathQA ($32.83 \rightarrow 33.53$), while maintaining strong alignment-sensitive performance on TruthfulQA and WSC273. These results demonstrate that performance-driven stochastic $\lambda$ adaptation is more robust than any static weighting scheme.

\paragraph{Benchmark-Specific Preference Sensitivity.}
Using the results reported in Table~\ref{tab:lambda-ablation}, we further analyze how different benchmarks respond to individual preference dimensions. Rather than focusing on overall performance aggregation, this analysis adopts a benchmark-centric perspective and examines which preference objectives are most aligned with specific evaluation tasks.

We observe that no single preference dimension dominates across all benchmarks. For example, ARC exhibits its strongest performance under the instruction-following–focused configuration, suggesting that precise adherence to task instructions plays a critical role in this benchmark. In contrast, TruthfulQA benefits more consistently from balanced or honesty-oriented preference mixtures, reflecting its sensitivity to factual correctness and avoidance of hallucination. Similarly, WSC273 shows improved performance under configurations that emphasize balanced reasoning and fluency, rather than any single isolated objective. These observations highlight that different benchmarks implicitly encode distinct preference structures

\section{Conclusions}
In this work, we revisit preference optimization for language model alignment from a multi-dimensional perspective. We identify that existing DPO-based methods implicitly assume a single fixed preference objective, which limits their expressive power when human judgments are structured, heterogeneous, and sometimes conflicting across dimensions. To address this limitation, we propose $\lambda$-weighted Listwise Direct Preference Optimization ($\lambda$-DPO), a unified framework that combines listwise supervision with flexible preference aggregation for multi-dimensional alignment. Extensive experiments across multiple model families and six widely used benchmarks demonstrate that $\lambda$-DPO consistently outperforms strong DPO-based baselines.

\section{Limitations}
While our method improves the flexibility and expressiveness of preference optimization, this work still has several limitations that warrant further investigation. Our evaluation primarily relies on accuracy over standard benchmarks as a proxy for alignment quality. Although these benchmarks capture important aspects such as reasoning ability, factual correctness, and instruction-following, they do not fully reflect the complexity of real-world deployment scenarios, including long-horizon interactions, dynamically changing user preferences, and safety-critical behaviors.
\bibliography{bib-final}

@article{brown2020language,
  title={Language models are few-shot learners},
  author={Brown, Tom and Mann, Benjamin and Ryder, Nick and Subbiah, Melanie and Kaplan, Jared D and Dhariwal, Prafulla and Neelakantan, Arvind and Shyam, Pranav and Sastry, Girish and Askell, Amanda and others},
  journal={Advances in neural information processing systems},
  volume={33},
  pages={1877--1901},
  year={2020}
}

@misc{touvron2023llamaopenefficientfoundation,
      title={LLaMA: Open and Efficient Foundation Language Models}, 
      author={Hugo Touvron and Thibaut Lavril and Gautier Izacard and Xavier Martinet and Marie-Anne Lachaux and Timothée Lacroix and Baptiste Rozière and Naman Goyal and Eric Hambro and Faisal Azhar and Aurelien Rodriguez and Armand Joulin and Edouard Grave and Guillaume Lample},
      year={2023},
      eprint={2302.13971},
      archivePrefix={arXiv},
      primaryClass={cs.CL},
      url={https://arxiv.org/abs/2302.13971}, 
}

@misc{openai2024gpt4technicalreport,
      title={GPT-4 Technical Report}, 
      author={OpenAI},
      year={2024},
      eprint={2303.08774},
      archivePrefix={arXiv},
      primaryClass={cs.CL},
      url={https://arxiv.org/abs/2303.08774}, 
}

@article{bai2022training,
  title={Training a helpful and harmless assistant with reinforcement learning from human feedback},
  author={Bai, Yuntao and Jones, Andy and Ndousse, Kamal and Askell, Amanda and Chen, Anna and DasSarma, Nova and Drain, Dawn and Fort, Stanislav and Ganguli, Deep and Henighan, Tom and others},
  journal={arXiv preprint arXiv:2204.05862},
  year={2022}
}

@article{ouyang2022training,
  title={Training language models to follow instructions with human feedback},
  author={Ouyang, Long and Wu, Jeffrey and Jiang, Xu and Almeida, Diogo and Wainwright, Carroll and Mishkin, Pamela and Zhang, Chong and Agarwal, Sandhini and Slama, Katarina and Ray, Alex and others},
  journal={Advances in neural information processing systems},
  volume={35},
  pages={27730--27744},
  year={2022}
}

@article{clark2018think,
  title={Think You Have Solved Question Answering? Try ARC, the AI2 Reasoning Challenge},
  author={Clark, Peter and others},
  journal={arXiv preprint arXiv:1803.05457},
  year={2018}
}

@misc{qwen2025qwen25technicalreport,
      title={Qwen2.5 Technical Report}, 
      author={Qwen},
      year={2025},
      eprint={2412.15115},
      archivePrefix={arXiv},
      primaryClass={cs.CL},
      url={https://arxiv.org/abs/2412.15115}, 
}

@inproceedings{lin-etal-2022-truthfulqa,
    title = "{T}ruthful{QA}: Measuring How Models Mimic Human Falsehoods",
    author = "Lin, Stephanie  and
      Hilton, Jacob  and
      Evans, Owain",
    editor = "Muresan, Smaranda  and
      Nakov, Preslav  and
      Villavicencio, Aline",
    booktitle = "Proceedings of the 60th Annual Meeting of the Association for Computational Linguistics (Volume 1: Long Papers)",
    month = may,
    year = "2022",
    address = "Dublin, Ireland",
    publisher = "Association for Computational Linguistics",
    url = "https://aclanthology.org/2022.acl-long.229/",
    doi = "10.18653/v1/2022.acl-long.229",
    pages = "3214--3252"
}

@article{levesque2012winograd,
  title={The Winograd Schema Challenge},
  author={Levesque, Hector and others},
  journal={KR},
  year={2012}
}

@inproceedings{
hendrycks2021measuring,
title={Measuring Massive Multitask Language Understanding},
author={Dan Hendrycks and Collin Burns and Steven Basart and Andy Zou and Mantas Mazeika and Dawn Song and Jacob Steinhardt},
booktitle={International Conference on Learning Representations},
year={2021},
url={https://openreview.net/forum?id=d7KBjmI3GmQ}
}

@article{christiano2017deep,
  title={Deep reinforcement learning from human preferences},
  author={Christiano, Paul F and Leike, Jan and Brown, Tom and Martic, Miljan and Legg, Shane and Amodei, Dario},
  journal={Advances in neural information processing systems},
  volume={30},
  year={2017}
}

@article{wu2021recursively,
  title={Recursively summarizing books with human feedback},
  author={Wu, Jeff and Ouyang, Long and Ziegler, Daniel M and Stiennon, Nisan and Lowe, Ryan and Leike, Jan and Christiano, Paul},
  journal={arXiv preprint arXiv:2109.10862},
  year={2021}
}

@inproceedings{korbak2023pretraining,
  title={Pretraining language models with human preferences},
  author={Korbak, Tomasz and Shi, Kejian and Chen, Angelica and Bhalerao, Rasika Vinayak and Buckley, Christopher and Phang, Jason and Bowman, Samuel R and Perez, Ethan},
  booktitle={International Conference on Machine Learning},
  pages={17506--17533},
  year={2023},
  organization={PMLR}
}

@article{bradley1952rank,
  title={Rank analysis of incomplete block designs: I. the method of paired comparisons},
  author={Bradley, Ralph Allan and Terry, Milton E},
  journal={Biometrika},
  volume={39},
  number={3/4},
  pages={324--345},
  year={1952},
  publisher={JSTOR}
}

@inproceedings{burges2005learning,
  title={Learning to rank using gradient descent},
  author={Burges, Chris and Shaked, Tal and Renshaw, Erin and Lazier, Ari and Deeds, Matt and Hamilton, Nicole and Hullender, Greg},
  booktitle={Proceedings of the 22nd international conference on Machine learning},
  pages={89--96},
  year={2005}
}

@article{cui2023ultrafeedback,
  title={Ultrafeedback: Boosting language models with high-quality feedback},
  author={Cui, Ganqu and Yuan, Lifan and Ding, Ning and Yao, Guanming and Zhu, Wei and Ni, Yuan and Xie, Guotong and Liu, Zhiyuan and Sun, Maosong},
  year={2023}
}

@misc{schulman2017proximalpolicyoptimizationalgorithms,
      title={Proximal Policy Optimization Algorithms}, 
      author={John Schulman and Filip Wolski and Prafulla Dhariwal and Alec Radford and Oleg Klimov},
      year={2017},
      eprint={1707.06347},
      archivePrefix={arXiv},
      primaryClass={cs.LG},
      url={https://arxiv.org/abs/1707.06347}, 
}

@article{yuan2023rrhf,
  title={Rrhf: Rank responses to align language models with human feedback},
  author={Yuan, Hongyi and Yuan, Zheng and Tan, Chuanqi and Wang, Wei and Huang, Songfang and Huang, Fei},
  journal={Advances in Neural Information Processing Systems},
  volume={36},
  pages={10935--10950},
  year={2023}
}

@article{rafailov2023direct,
  title={Direct preference optimization: Your language model is secretly a reward model},
  author={Rafailov, Rafael and Sharma, Archit and Mitchell, Eric and Manning, Christopher D and Ermon, Stefano and Finn, Chelsea},
  journal={Advances in neural information processing systems},
  volume={36},
  pages={53728--53741},
  year={2023}
}

@misc{lai2024stepdpostepwisepreferenceoptimization,
      title={Step-DPO: Step-wise Preference Optimization for Long-chain Reasoning of LLMs}, 
      author={Xin Lai and Zhuotao Tian and Yukang Chen and Senqiao Yang and Xiangru Peng and Jiaya Jia},
      year={2024},
      eprint={2406.18629},
      archivePrefix={arXiv},
      primaryClass={cs.LG},
      url={https://arxiv.org/abs/2406.18629}, 
}

@inproceedings{amini-etal-2019-mathqa,
    title = "{M}ath{QA}: Towards Interpretable Math Word Problem Solving with Operation-Based Formalisms",
    author = "Amini, Aida  and
      Gabriel, Saadia  and
      Lin, Shanchuan  and
      Koncel-Kedziorski, Rik  and
      Choi, Yejin  and
      Hajishirzi, Hannaneh",
    editor = "Burstein, Jill  and
      Doran, Christy  and
      Solorio, Thamar",
    booktitle = "Proceedings of the 2019 Conference of the North {A}merican Chapter of the Association for Computational Linguistics: Human Language Technologies, Volume 1 (Long and Short Papers)",
    month = jun,
    year = "2019",
    address = "Minneapolis, Minnesota",
    publisher = "Association for Computational Linguistics",
    url = "https://aclanthology.org/N19-1245/",
    doi = "10.18653/v1/N19-1245",
    pages = "2357--2367"
}

@misc{bai2025betteroptimizationlistwisepreference,
      title={Towards Better Optimization For Listwise Preference in Diffusion Models}, 
      author={Jiamu Bai and Xin Yu and Meilong Xu and Weitao Lu and Xin Pan and Kiwan Maeng and Daniel Kifer and Jian Wang and Yu Wang},
      year={2025},
      eprint={2510.01540},
      archivePrefix={arXiv},
      primaryClass={cs.CV},
      url={https://arxiv.org/abs/2510.01540}, 
}

@inproceedings{zellers-etal-2019-hellaswag,
    title = "{H}ella{S}wag: Can a Machine Really Finish Your Sentence?",
    author = "Zellers, Rowan  and
      Holtzman, Ari  and
      Bisk, Yonatan  and
      Farhadi, Ali  and
      Choi, Yejin",
    editor = "Korhonen, Anna  and
      Traum, David  and
      M{\`a}rquez, Llu{\'i}s",
    booktitle = "Proceedings of the 57th Annual Meeting of the Association for Computational Linguistics",
    month = jul,
    year = "2019",
    address = "Florence, Italy",
    publisher = "Association for Computational Linguistics",
    url = "https://aclanthology.org/P19-1472/",
    doi = "10.18653/v1/P19-1472",
    pages = "4791--4800"
}

@inproceedings{pesaran-zadeh-etal-2025-lpoi,
    title = "{LPOI}: Listwise Preference Optimization for Vision Language Models",
    author = "Pesaran Zadeh, Fatemeh  and
      Oh, Yoojin  and
      Kim, Gunhee",
    editor = "Che, Wanxiang  and
      Nabende, Joyce  and
      Shutova, Ekaterina  and
      Pilehvar, Mohammad Taher",
    booktitle = "Proceedings of the 63rd Annual Meeting of the Association for Computational Linguistics (Volume 1: Long Papers)",
    month = jul,
    year = "2025",
    address = "Vienna, Austria",
    publisher = "Association for Computational Linguistics",
    url = "https://aclanthology.org/2025.acl-long.1302/",
    doi = "10.18653/v1/2025.acl-long.1302",
    pages = "26830--26844",
    ISBN = "979-8-89176-251-0"
}

@inproceedings{zhao-etal-2025-permutative,
    title = "Permutative Preference Alignment from Listwise Ranking of Human Judgments",
    author = "Zhao, Yang  and
      Wang, Yixin  and
      Yin, Mingzhang",
    editor = "Christodoulopoulos, Christos  and
      Chakraborty, Tanmoy  and
      Rose, Carolyn  and
      Peng, Violet",
    booktitle = "Proceedings of the 2025 Conference on Empirical Methods in Natural Language Processing",
    month = nov,
    year = "2025",
    address = "Suzhou, China",
    publisher = "Association for Computational Linguistics",
    url = "https://aclanthology.org/2025.emnlp-main.17/",
    doi = "10.18653/v1/2025.emnlp-main.17",
    pages = "310--334",
    ISBN = "979-8-89176-332-6"
}

@inproceedings{
pang2024iterative,
title={Iterative Reasoning Preference Optimization},
author={Richard Yuanzhe Pang and Weizhe Yuan and He He and Kyunghyun Cho and Sainbayar Sukhbaatar and Jason E Weston},
booktitle={The Thirty-eighth Annual Conference on Neural Information Processing Systems},
year={2024},
url={https://openreview.net/forum?id=4XIKfvNYvx}
}

@inproceedings{
tu2025enhancing,
title={Enhancing {LLM} Reasoning with Iterative {DPO}: A Comprehensive Empirical Investigation},
author={Songjun Tu and Jiahao Lin and Xiangyu Tian and Qichao Zhang and Linjing Li and Yuqian Fu and Nan Xu and Wei He and Xiangyuan Lan and Dongmei Jiang and Dongbin Zhao},
booktitle={Second Conference on Language Modeling},
year={2025},
url={https://openreview.net/forum?id=OgWh4J7bkT}
}

@article{meng2024simpo,
  title={Simpo: Simple preference optimization with a reference-free reward},
  author={Meng, Yu and Xia, Mengzhou and Chen, Danqi},
  journal={Advances in Neural Information Processing Systems},
  volume={37},
  pages={124198--124235},
  year={2024}
}

@inproceedings{ shi2025the, title={The Crucial Role of Samplers in Online Direct Preference Optimization}, author={Ruizhe Shi and Runlong Zhou and Simon Shaolei Du}, booktitle={The Thirteenth International Conference on Learning Representations}, year={2025}, url={https://openreview.net/forum?id=F6z3utfcYw} }

@misc{xiao2025comprehensivesurveydirectpreference,
      title={A Comprehensive Survey of Direct Preference Optimization: Datasets, Theories, Variants, and Applications}, 
      author={Wenyi Xiao and Zechuan Wang and Leilei Gan and Shuai Zhao and Zongrui Li and Ruirui Lei and Wanggui He and Luu Anh Tuan and Long Chen and Hao Jiang and Zhou Zhao and Fei Wu},
      year={2025},
      eprint={2410.15595},
      archivePrefix={arXiv},
      primaryClass={cs.AI},
      url={https://arxiv.org/abs/2410.15595}, 
}

@inproceedings{xu-etal-2025-full,
    title = "Full-Step-{DPO}: Self-Supervised Preference Optimization with Step-wise Rewards for Mathematical Reasoning",
    author = "Xu, Huimin  and
      Mao, Xin  and
      Li, Feng-Lin  and
      Wu, Xiaobao  and
      Chen, Wang  and
      Zhang, Wei  and
      Luu, Anh Tuan",
    editor = "Che, Wanxiang  and
      Nabende, Joyce  and
      Shutova, Ekaterina  and
      Pilehvar, Mohammad Taher",
    booktitle = "Findings of the Association for Computational Linguistics: ACL 2025",
    month = jul,
    year = "2025",
    address = "Vienna, Austria",
    publisher = "Association for Computational Linguistics",
    url = "https://aclanthology.org/2025.findings-acl.1249/",
    doi = "10.18653/v1/2025.findings-acl.1249",
    pages = "24343--24356",
    ISBN = "979-8-89176-256-5"
}

@inproceedings{park-etal-2024-disentangling,
    title = "Disentangling Length from Quality in Direct Preference Optimization",
    author = "Park, Ryan  and
      Rafailov, Rafael  and
      Ermon, Stefano  and
      Finn, Chelsea",
    editor = "Ku, Lun-Wei  and
      Martins, Andre  and
      Srikumar, Vivek",
    booktitle = "Findings of the Association for Computational Linguistics: ACL 2024",
    month = aug,
    year = "2024",
    address = "Bangkok, Thailand",
    publisher = "Association for Computational Linguistics",
    url = "https://aclanthology.org/2024.findings-acl.297/",
    doi = "10.18653/v1/2024.findings-acl.297",
    pages = "4998--5017"
}

@misc{shao2024deepseekmathpushinglimitsmathematical,
      title={DeepSeekMath: Pushing the Limits of Mathematical Reasoning in Open Language Models}, 
      author={Zhihong Shao and Peiyi Wang and Qihao Zhu and Runxin Xu and Junxiao Song and Xiao Bi and Haowei Zhang and Mingchuan Zhang and Y. K. Li and Y. Wu and Daya Guo},
      year={2024},
      eprint={2402.03300},
      archivePrefix={arXiv},
      primaryClass={cs.CL},
      url={https://arxiv.org/abs/2402.03300}, 
}

@misc{sharifnassab2024softpreferenceoptimizationaligning,
      title={Soft Preference Optimization: Aligning Language Models to Expert Distributions}, 
      author={Arsalan Sharifnassab and Saber Salehkaleybar and Sina Ghiassian and Surya Kanoria and Dale Schuurmans},
      year={2024},
      eprint={2405.00747},
      archivePrefix={arXiv},
      primaryClass={cs.LG},
      url={https://arxiv.org/abs/2405.00747}, 
}

@misc{hu2025reinforcestabilizingcriticfreepolicy,
      title={REINFORCE++: Stabilizing Critic-Free Policy Optimization with Global Advantage Normalization}, 
      author={Jian Hu and Jason Klein Liu and Haotian Xu and Wei Shen},
      year={2025},
      eprint={2501.03262},
      archivePrefix={arXiv},
      primaryClass={cs.CL},
      url={https://arxiv.org/abs/2501.03262}, 
}

@article{sutton1999policy,
  title={Policy gradient methods for reinforcement learning with function approximation},
  author={Sutton, Richard S and McAllester, David and Singh, Satinder and Mansour, Yishay},
  journal={Advances in neural information processing systems},
  volume={12},
  year={1999}
}

@inproceedings{liu-etal-2025-lipo,
    title = "{L}i{PO}: Listwise Preference Optimization through Learning-to-Rank",
    author = "Liu, Tianqi  and
      Qin, Zhen  and
      Wu, Junru  and
      Shen, Jiaming  and
      Khalman, Misha  and
      Joshi, Rishabh  and
      Zhao, Yao  and
      Saleh, Mohammad  and
      Baumgartner, Simon  and
      Liu, Jialu  and
      Liu, Peter J  and
      Wang, Xuanhui",
    editor = "Chiruzzo, Luis  and
      Ritter, Alan  and
      Wang, Lu",
    booktitle = "Proceedings of the 2025 Conference of the Nations of the Americas Chapter of the Association for Computational Linguistics: Human Language Technologies (Volume 1: Long Papers)",
    month = apr,
    year = "2025",
    address = "Albuquerque, New Mexico",
    publisher = "Association for Computational Linguistics",
    url = "https://aclanthology.org/2025.naacl-long.121/",
    doi = "10.18653/v1/2025.naacl-long.121",
    pages = "2404--2420",
    ISBN = "979-8-89176-189-6"
}

@inproceedings{
xu2024contrastive,
title={Contrastive Preference Optimization: Pushing the Boundaries of {LLM} Performance in Machine Translation},
author={Haoran Xu and Amr Sharaf and Yunmo Chen and Weiting Tan and Lingfeng Shen and Benjamin Van Durme and Kenton Murray and Young Jin Kim},
booktitle={Forty-first International Conference on Machine Learning},
year={2024},
url={https://openreview.net/forum?id=51iwkioZpn}
}

@inproceedings{azar2024general,
  title={A general theoretical paradigm to understand learning from human preferences},
  author={Azar, Mohammad Gheshlaghi and Guo, Zhaohan Daniel and Piot, Bilal and Munos, Remi and Rowland, Mark and Valko, Michal and Calandriello, Daniele},
  booktitle={International Conference on Artificial Intelligence and Statistics},
  pages={4447--4455},
  year={2024},
  organization={PMLR}
}

@inproceedings{hong-etal-2024-orpo,
    title = "{ORPO}: Monolithic Preference Optimization without Reference Model",
    author = "Hong, Jiwoo  and
      Lee, Noah  and
      Thorne, James",
    editor = "Al-Onaizan, Yaser  and
      Bansal, Mohit  and
      Chen, Yun-Nung",
    booktitle = "Proceedings of the 2024 Conference on Empirical Methods in Natural Language Processing",
    month = nov,
    year = "2024",
    address = "Miami, Florida, USA",
    publisher = "Association for Computational Linguistics",
    url = "https://aclanthology.org/2024.emnlp-main.626/",
    doi = "10.18653/v1/2024.emnlp-main.626",
    pages = "11170--11189"
}


\appendix

\section{Mathematical Derivations}
\label{sec:derivation}

\subsection{Deriving the Final Training Objective}
We begin by defining the base reward model used in DPO. Let $\pi_\theta(y|x)$ denote the current policy and $\pi_{\text{ref}}(y|x)$ a fixed reference policy. Given input $x$ and candidate outputs $\{y_1, \ldots, y_N\}$, the reward is defined via KL divergence as:
\begin{equation}
r_\theta(x, y) = \beta \log \left( \frac{\pi_\theta(y|x)}{\pi_{\text{ref}}(y|x)} \right) + \beta \log Z(x)
\end{equation}
where $Z(x)$ is a normalizing constant ensuring scale-invariance:
\begin{equation}
Z(x) = \sum_{j=1}^N \left( \frac{\pi_\theta(y_j|x)}{\pi_{\text{ref}}(y_j|x)} \right)
\end{equation}

The predicted distribution over completions is then:
\begin{equation}
P_\theta(y_i|x) = \frac{\exp(r_\theta(x, y_i))}{\sum_{j=1}^N \exp(r_\theta(x, y_j))}
\end{equation}

Given a human-provided listwise preference distribution $p^*_i$, we minimize the cross-entropy between $p^*_i$ and $P_\theta$:
\begin{align}
\mathcal{L}^{\text{DPO}}(\theta) = & \mathbb{E}_{(x, \{y_i\}) \sim \mathcal{D}} \bigg[ - \sum_{i=1}^N p^*_i \nonumber \\
& \cdot \log \left( \frac{\exp(r_\theta(x, y_i))}{\sum_{j=1}^N \exp(r_\theta(x, y_j))} \right) \bigg]
\end{align}

Substituting the reward expression and simplifying:
\begin{center}
\scalebox{0.875}{$
\begin{aligned}
& \mathcal{L}^{\text{DPO}}(\theta)\\
= &  \mathbb{E}_{(x, \{y_i\}) \sim \mathcal{D}} \left[ - \sum_{i=1}^N p^*_i \cdot \log \left( \frac{ \left( \frac{\pi_\theta(y_i|x)}{\pi_{\text{ref}}(y_i|x)} \right)^\beta Z(x)^\beta }{ \sum_{j=1}^N \left( \frac{\pi_\theta(y_j|x)}{\pi_{\text{ref}}(y_j|x)} \right)^\beta Z(x)^\beta } \right) \right] \\
= & \mathbb{E}_{(x, \{y_i\}) \sim \mathcal{D}} \left[ - \sum_{i=1}^N p^*_i \cdot \log \left( \frac{ \left( \frac{\pi_\theta(y_i|x)}{\pi_{\text{ref}}(y_i|x)} \right)^\beta }{ \sum_{j=1}^N \left( \frac{\pi_\theta(y_j|x)}{\pi_{\text{ref}}(y_j|x)} \right)^\beta } \right) \right]
\end{aligned}
$}
\end{center}

This becomes the canonical DPO loss. To extend this to multiple preference dimensions, we assume there are $m$ reward components indexed by $k$, each with its own preference distribution $p_i^{*(k)}$:
\begin{align}
\mathcal{L}^{(k)}(\theta) = & \mathbb{E}_{(x, \{y_i\}) \sim \mathcal{D}} \bigg[ - \sum_{i=1}^N p_i^{*(k)} \nonumber \\
& \cdot \log \left( \frac{\left( \frac{ \pi_\theta(y_i|x) }{ \pi_{\text{ref}}(y_i|x) } \right)^\beta }{ \sum_{j=1}^N \left( \frac{ \pi_\theta(y_j|x) }{ \pi_{\text{ref}}(y_j|x) } \right)^\beta } \right) \bigg]
\end{align}

We then define a user-controllable or sampled weight vector $\lambda \in \Delta^m$ (the $m$-dimensional probability simplex: $\lambda_k \ge 0$, $\sum_k \lambda_k = 1$), and define the aggregated loss:
\begin{equation}
\mathcal{L}^\lambda(\theta) = \sum_{k=1}^{m} \lambda_k \cdot \mathcal{L}^{(k)}(\theta)
\end{equation}

To enable generalization over arbitrary preferences, we optimize the expected loss over $\lambda \sim \text{Unif}(\Delta^m)$:
\begin{align}
\mathcal{L}_{\text{final}}(\theta) &= \mathbb{E}_{\lambda \sim \text{Unif}(\Delta^m)} \left[ \mathcal{L}^\lambda(\theta) \right] \\
& = \mathbb{E}_{\lambda \sim \text{Unif}(\Delta^m)} \left[ \sum_{k=1}^m \lambda_k \cdot \mathcal{L}^{(k)}(\theta) \right]
\end{align}

\subsection{From Final Objective to Lambda-DPO Form}
To convert Eq. (25) to a form that operates on distributions directly, we change the order of summation. Recall:
\begin{equation}
\mathcal{L}^{(k)}(\theta) = \mathbb{E}_{(x, \{y_i\}) \sim \mathcal{D}} \left[ - \sum_{i=1}^N p_i^{*(k)} \cdot \log P_\theta(y_i \mid x) \right]
\end{equation}
So the aggregated expected loss becomes:

\begin{align}
\mathcal{L}_{\text{final}}(\theta) &= \mathbb{E}_{\lambda \sim \text{Unif}(\Delta^m)} \Bigg\{ \sum_{k=1}^m \lambda_k \nonumber \\ 
& \cdot \mathbb{E}_{(x, \{y_i\})} \left[ - \sum_{i=1}^N p_i^{*(k)} \log P_\theta(y_i|x) \right] \Bigg\} \\
&= \mathbb{E}_{(x, \{y_i\}), \lambda} \left[ - \sum_{i=1}^N \sum_{k=1}^m \lambda_k p_i^{*(k)} \log P_\theta(y_i|x) \right]
\end{align}

We now define the aggregated distribution over outputs:
\begin{equation}
p^\lambda(y_i|x) = \sum_{k=1}^m \lambda_k p^{*(k)}(y_i|x)
\end{equation}

So the objective becomes:
\begin{align}
& \mathcal{L}_{\lambda\text{-DPO}}(\theta) \nonumber \\
& = - \mathbb{E}_{(x, \{y_i\}), \lambda} \left[ \sum_{i=1}^{N} p^\lambda(y_i \mid x) \cdot \log P_\theta(y_i \mid x) \right]
\end{align}

Thus, by switching the summation order (i.e., aggregating over preference weights before computing loss), we arrive at an equivalent but more compact objective that is easier to optimize and sample in practice.

\subsection{A.3 Gradient of Lambda-DPO Loss}
We compute the gradient of $\mathcal{L}_{\lambda\text{-DPO}}$:
\begin{align}
\nabla_\theta \mathcal{L}_{\lambda\text{-DPO}} & \nonumber \\
= - \mathbb{E}_{x, \lambda} & \left[ \sum_{i=1}^N p^\lambda(y_i \mid x) \cdot \nabla_\theta \log P_\theta(y_i \mid x) \right] \\
= -\mathbb{E}_{x, \lambda} & \bigg[ \sum_i \left( p^\lambda(y_i \mid x) - P_\theta(y_i \mid x) \right) \nonumber \\ 
& \qquad \quad \cdot \nabla_\theta \log \pi_\theta(y_i \mid x) \bigg]
\end{align}
where we use:
\begin{equation}
\nabla_\theta P_\theta(y_i|x) = P_\theta(y_i|x) \cdot \nabla_\theta \log \pi_\theta(y_i|x)
\end{equation}
This completes the derivation of the Lambda-weighted DPO objective and its gradient form.

\section{Benchmark Statistics}
\label{sec: bench}
We report average accuracy across the following six established datasets:

\begin{itemize}
  \item \textbf{MMLU}: 57 subjects, 15K questions assessing academic knowledge~\cite{hendrycks2021measuring}.
  \item \textbf{ARC}: 1.1K grade-school science questions requiring reasoning~\cite{clark2018think}.
  \item \textbf{HellaSwag}: 10K commonsense sentence completions with distractors~\cite{zellers-etal-2019-hellaswag}.
  \item \textbf{TruthfulQA}: 817 questions testing factuality and hallucination~\cite{lin-etal-2022-truthfulqa}.
  \item \textbf{Winograd}: 273 examples probing coreference and syntax~\cite{levesque2012winograd}.
  \item \textbf{MathQA}: 37k Math reasoning problems~\cite{amini-etal-2019-mathqa}.
\end{itemize}

\section{Performance-Driven Lambda Scheduling via Polynomial Modeling}
\label{sec:scheduler}

We introduce a general and flexible method for \emph{performance-driven lambda scheduling}, which constructs a probability distribution over preference vectors $\lambda \in \Delta^d$ based on observed alignment performance. Each supervised training point $(\lambda^{(i)}, y^{(i)})$ consists of a preference configuration $\lambda^{(i)}$ and a corresponding scalar performance metric $y^{(i)}$ (e.g., average benchmark accuracy). The scheduling process unfolds in four steps:

\subsection{Polynomial Feature Mapping}

We begin by mapping each preference vector $\lambda$ to a high-dimensional feature space using a polynomial embedding. Specifically, we define a degree-$p$ polynomial feature map $\phi_p : \mathbb{R}^d \rightarrow \mathbb{R}^N$ that enumerates all monomials up to degree $p$:
\begin{equation}
\phi_p(\lambda) = \left[1, \lambda_1, \dots, \lambda_d, \lambda_1^2, \lambda_1 \lambda_2, \dots, \lambda_d^p\right]^\top,
\end{equation}
where the total number of features is given by:
\begin{equation}
N = \binom{d + p}{p}.
\end{equation}
This expansion enables expressive yet tractable modeling of non-linear relationships between $\lambda$ vectors and alignment performance.

\subsection{Polynomial Regression}

Using the training set $\{(\lambda^{(i)}, y^{(i)})\}_{i=1}^n$, we fit a least-squares regression model in the lifted feature space to learn the performance landscape:
\begin{equation}
\min_{\mathbf{w} \in \mathbb{R}^N} \sum_{i=1}^{n} \left( \mathbf{w}^\top \phi_p(\lambda^{(i)}) - y^{(i)} \right)^2.
\end{equation}
The resulting predictive function
\begin{equation}
f(\lambda) = \mathbf{w}^\top \phi_p(\lambda)
\end{equation}
approximates the expected performance of any candidate $\lambda$ on downstream alignment tasks.

\subsection{Grid Construction over the Simplex}

To enable tractable sampling from the $\lambda$ space, we construct a discrete candidate set $\{\lambda^{(j)}\}_{j=1}^k \subset \Delta^d$ via one of the following methods:
\begin{itemize}
    \item \textbf{Simplex discretization:} generate all combinations with replacement and normalize;
    \item \textbf{Dirichlet sampling:} draw $\lambda^{(j)} \sim \mathrm{Dir}(\boldsymbol{\alpha})$, where $\boldsymbol{\alpha} = \mathbf{1}$ yields a uniform prior over the simplex.
\end{itemize}

\subsection{Softmax-Based Scheduling Distribution}

We convert the predicted scores $s_j = f(\lambda^{(j)})$ into a probability distribution over the candidate $\lambda$ vectors using a softmax transformation:
\begin{equation}
p(\lambda^{(j)}) = \frac{\exp(\tau \cdot f(\lambda^{(j)}))}{\sum_{l=1}^{k} \exp(\tau \cdot f(\lambda^{(l)}))},
\end{equation}
where the temperature parameter $\tau > 0$ controls the exploration–exploitation trade-off.

\subsection{Empirical Instantiation with Five Training Points}

To validate the polynomial scheduling framework, we use five observed training points—four unimodal $\lambda$ vectors and one uniform configuration—each paired with an empirical alignment accuracy:
\begin{align*}
\lambda^{(1)} &= [1, 0, 0, 0], \quad y^{(1)} = 0.4563 \\
\lambda^{(2)} &= [0, 1, 0, 0], \quad y^{(2)} = 0.4561 \\
\lambda^{(3)} &= [0, 0, 1, 0], \quad y^{(3)} = 0.4578 \\
\lambda^{(4)} &= [0, 0, 0, 1], \quad y^{(4)} = 0.4553 \\
\lambda^{(5)} &= [0.25, 0.25, 0.25, 0.25], \quad y^{(5)} = 0.4623
\end{align*}

We set $p=2$, yielding $N = \binom{4+2}{2} = 15$ features. Using gradient descent to minimize squared error, we obtain:
\begin{align*}
f(\lambda) &= 0.36
+ 0.09 \lambda_1 + 0.09 \lambda_2 + 0.09 \lambda_3 + 0.09 \lambda_4 \\
&+ 0.0061 \lambda_1^2
+ 0.0280 \lambda_1 \lambda_2
+ 0.0280 \lambda_1 \lambda_3 \\
&+ 0.0280 \lambda_1 \lambda_4 
+ 0.0060 \lambda_2^2
+ 0.0280 \lambda_2 \lambda_3 \\
&+ 0.0280 \lambda_2 \lambda_4 
+ 0.0068 \lambda_3^2 \\
&+ 0.0280 \lambda_3 \lambda_4
+ 0.0056 \lambda_4^2.
\end{align*}

\subsection{Lambda Sampling from Learned Scheduler}

We sample $k=10$ $\lambda$ vectors from Dirichlet$(1,1,1,1)$ and compute predicted scores $f(\lambda^{(j)})$, converting them via softmax ($\tau=100$):
\[
p(\lambda^{(j)}) = \frac{\exp(\tau \cdot f(\lambda^{(j)}))}{\sum_{l=1}^{10} \exp(\tau \cdot f(\lambda^{(l)}) )}
\]

\begin{table}[t]
\centering
\begin{tabular}{rrrrrr}
\toprule
$\lambda_1$ & $\lambda_2$ & $\lambda_3$ & $\lambda_4$ & $f(\lambda)$ & $p(\lambda)$ \\
\midrule
0.212 & 0.334 & 0.245 & 0.209 & 0.462 & 0.108 \\
0.126 & 0.237 & 0.131 & 0.507 & 0.461 & 0.098 \\
0.542 & 0.079 & 0.256 & 0.123 & 0.461 & 0.099 \\
0.233 & 0.721 & 0.020 & 0.025 & 0.459 & 0.082 \\
0.004 & 0.334 & 0.281 & 0.381 & 0.462 & 0.101 \\
0.320 & 0.069 & 0.513 & 0.099 & 0.462 & 0.104 \\
0.236 & 0.155 & 0.358 & 0.251 & 0.462 & 0.108 \\
0.141 & 0.110 & 0.701 & 0.048 & 0.461 & 0.095 \\
0.070 & 0.198 & 0.388 & 0.344 & 0.462 & 0.104 \\
0.090 & 0.139 & 0.306 & 0.465 & 0.462 & 0.100 \\
\bottomrule
\end{tabular}
\caption{Sampled $\lambda$ vectors, their predicted scores $f(\lambda)$, and softmax sampling probabilities $p(\lambda)$ at temperature $\tau = 100$.}
\label{tab:scheduler-sampling}
\end{table}

\subsection{Scheduler vs. Uniform Comparison}

We compare our learned scheduler with a uniform $\lambda = [\frac{1}{4}, \frac{1}{4}, \frac{1}{4}, \frac{1}{4}]$ baseline using Llama 3.2-1B Instruct. The learned scheduler achieves slightly better average accuracy (\%) (46.82 vs. 46.23), while additionally introducing stochastic $\lambda$ sampling as a regularization mechanism, improving robustness.

\end{document}